# The Challenges and Opportunities of Human-Centered AI for Trustworthy Robots and Autonomous Systems

Hongmei He, *Senior Member, IEEE*, John Gray, *Member, IEEE*,

Angelo Cangelosi, *Senior Member, IEEE,* Qinggang Meng, *Senior Member*, *IEEE*,

T.M. McGinnity, *Senior Member*, *IEEE,* Jörn Mehnen, *Member, IEEE*

*Abstract—* **The trustworthiness of Robots and Autonomous Systems (RAS) has gained a prominent position on many research agendas towards fully autonomous systems. This research systematically explores, for the first time, the key facets of human-centered AI (HAI) for trustworthy RAS. In this article, five key properties of a trustworthy RAS initially have been identified. RAS must be (i) safe in any uncertain and dynamic surrounding environments; (ii) secure, thus protecting itself from any cyber-threats; (iii) healthy with fault tolerance; (iv) trusted and easy to use to allow effective human-machine interaction (HMI), and (v) compliant with the law and ethical expectations. Then, the challenges in implementing trustworthy autonomous system are analytically reviewed, in respects of the five key properties, and the roles of AI technologies have been explored to ensure the trustiness of RAS with respects to safety, security, health and HMI, while reflecting the requirements of ethics in the design of RAS. While applications of RAS have mainly focused on performance and productivity, the risks posed by advanced AI in RAS have not received sufficient scientific attention. Hence, a new acceptance model of RAS is provided, as a framework for requirements to human-centered AI and for implementing trustworthy RAS by *design*. This approach promotes human-level intelligence to augments human's capacity while focusing on contributions to humanity.**

## I. INTRODUCTION

Robots and Autonomous Systems (RAS), equipped with Artificial Intelligence (AI) technologies, including Machine Learning (ML), are attempting to mimic the adaptive and smart capabilities of human problem solving [1]. RAS allow for creating smart and safe work environments where humans are relieved from the burden of arduous, repetitive or dangerous tasks. RAS can also help where super-human quick and precise actions are of the essence. For example, driverless cars equipped with intelligent safe road assist systems can greatly reduce the frequency of road accidents, medical robots with smart augmented reality technology enhance the performance of intricate surgeries, and intelligent autopilots perform delicate docking manoeuvres.

The vast benefits of RAS have made this technology popular in many application domains such as aerospace, transport, manufacturing, agriculture, social healthcare, and extreme environments. The spectrum of RAS applications spans e.g. robotics, autonomous vehicles, unmanned aerial vehicles (UAV), autonomous trading systems, self-managing telecommunication networks, smart factories, and infrastructure.

The Internet of Things (IoT) delivers new value by connecting people, processes and data. Sensing and data analysis technologies in IoT are giving robots a wider situational awareness that leads to better task execution. The concept of the Internet of Robotic Things (IoRT), raised by ABI research [2], introduces robots into the IoT application domains, creating harmonic collaboration between human, machines, and the physical world. Hence, IoRT technology extends the application scope of RAS and makes it an extremely powerful tool.

Even with the best technology however, it is impossible to anticipate all potential challenging situations that a RAS in real-world may experience. Advanced RAS must behave robustly and safely in any critical situation. Trust is built on predictability and understanding. Therefore, Trustworthy RAS (TRAS) by design must be the starting point to ensure progress towards trustworthy, fully autonomous systems.

As expressed by the Gartner Hype-Cycle in AI 2020 [3], the trustworthiness of AI is the top AI challenge today. People tend to initially view new technologies overly optimistically and may ignore or be initially unaware of the potential risks. Curiosity towards a new technology may lead to initial

H. He is with the School of Computer Science and Informatics, De Montfort University, Leicester, UK, LE1 9BH (e-mail: h.he@dmu.ac.uk).

J. Gray is with the Department of Electronics and Electrical Engineering, the University of Manchester, Manchester, UK. M13 9PL (e-mail: john.gray-2@manchester.ac.uk).

A. Cangelosi is the Department of Computer Science, the University of Manchester, Manchester, UK, M13 9PL (e-mail: angelo.cangelosi@manchester.ac.uk).

Q. Meng is with the Department of Computer Science, Loughborough University, Loughborough, UK, LE11 3TU (e-mail: q.meng@lboro.ac.uk).

T.M. McGinnity is with the Department of Computer Science, Nottingham Trent University, Nottingham, UK, NG1 4FQ (e-mail: martin.mcginnity@ntu.ac.uk) and also with the Intelligent Systems Research Centre, Ulster University (tm.mcginnity@ulster.ac.uk)

J. Mehnen is with Design, Manufacturing and Eng. Management, University of Strathclyde, Glasgow, UK, G1 1XQ (e-mail: jorn.mehnen@strath.ac.uk).

acceptance of untrustworthy products and services. However, as time and technology progresses, one gradually comes to realise that these systems must be trustworthy to reach the plateau of productivity. With regards to autonomous systems, hard lessons have been learnt. For example, a self-driving car hit a lady in Arizona in 2018 in a fatal accident [4]; hijacking a car on a motorway proved successfully that security flaws exist in a modern remote accessible vehicles [5]; two airplane accidents that costed 346 human lives were caused by failures of the onboard intelligent aviation systems [6]. Some intelligent robotic systems have catastrophically failed in situations where they were supposed to provide a high level of safety. Such accidents damage the public trust in advanced RAS technologies.

Implementing trustworthiness in RAS is not just a technical question. The examples above demonstrate that trustworthy RAS requires addressing ethical, societal, regulatory and educational challenges as well as technological issues.[7]. One of the most common ethical examples is that autonomous systems might face the so-called "trolley problem" [8], which involves stylized ethical dilemmas of whether to sacrifice one person to save a larger number. While some are regarding the incorporation of trustworthiness in RAS only as an additional cost burden, in September 2020 the European Parliament estimated in its latest European Value Added Assessment that a joint EU approach to ethical aspects of AI can add an extra €294 billion in GDP and 4.6 million jobs in the EU by 2030 [9].

In this paper, we address the important properties in technologies and ethics for implementing trustworthy RAS. Especially, we explore the challenges and opportunities of AI in implementing trustworthy RAS in respect of the identified properties in technologies. Also, we provide an acceptance model of human to RAS and highlight human-centered AI on the path towards fully autonomous systems. The UKRI is investing £33M in a Trustworthy Autonomous Systems (TAS) program [10]. This research completely aligns with the UKRI TAS program.

The paper is organized as follows. Section II will identify important factors that affect the trustworthiness of RAS; Section III will address the challenges of implementing trustworthy RAS in respects of the five key properties; Section IV will explore the roles of AI in implementing trustworthy RAS; Section V proposes the acceptance model of human to RAS and addresses the importance of human-centered AI, and finally, Section VI concludes the work.

## II. Key Factors That Affect TRAS

The growing concern over the need for trustworthy RAS has led to some initial international efforts to develop approaches to ensure and enhance trustworthiness. In the USA, the National Institute of Standards and Technology (NIST) has developed a framework encapsulating the key aspects of RAS, specifically in the context of Cyber Physical Systems (CPS) [11], as shown in Fig. 1. In the NIST model, we added a red arrow from cyber space to physical space, indicating that the cybersecurity of the physical layer could inherit the security issues from IT infrastructure, and directly affect the decisions and actions of the physical layer.

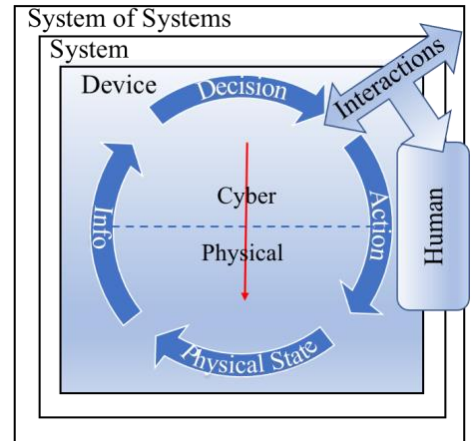

Fig. 1. The NIST CPS Framework Release 1.0 [11]

The trustworthiness of RAS is an important global issue, as global technological companies are at the forefront of AI and RAS developments. It is logical therefore that a more coordinated, systematic international effort must ensure that humans can develop and assess trust in RAS. In the following, we summarise critical factors of RAS, which need thorough consideration to convince a currently sceptical public of trustworthy RAS.

A robot or autonomous system will only be considered trustworthy if it can demonstrate advanced, reliable, safe performance in its core functionality. For most RAS, such characteristics begin with the system's sensing, and data collection modalities, upon which are built the perception, decision making, communication and actions of the entity. Logicality, dependability, correctness and transparency of decision-making, performance and overall quality are essential requirements of autonomous systems upon which trust is built.

*Safety* is an essential requirement for all kinds of RAS. Different applications of RAS may require different levels of safety. Clearly for applications, such as for the aerospace industry, the safety of passengers and crew is an absolutely fundamental criterion; for automobile industry, safety of vehicle users, other road users and assets in surrounding environments are of vital importance. To ensure the safety and monitor the surrounding environments, modern automobiles are increasingly incorporating many different technologies, including cameras, GPS, radar, LiDAR, and other types of sensors as well as on-board computers. Conversely safety of an autonomous domestic appliance, such as a robotic hoover, is important, but a failure would have less catastrophic effects. Safety is directly associated with RAS reliability, which is an important factor to be considered when a human selects a RAS.

*Security* is now of major concern to the public. This topic has accelerated towards the top of user concerns, as a result of high-visibility data breaches in major companies. However, data breaches are only one aspect of the security of RAS. Possibly more challenging is the security for cyber-physical systems, where cyber-attacks not only threaten information leakage, but also directly take control of the entity. Security is particularly important for the automobile sector but has implications for the service sector as well. In modern vehicles, the driver's role is increasingly being transferred from being active in control to that of a passive passenger. Exploiting data secured from engine management systems, a plethora of sensors connect to inform the vehicle's autonomous system. As with any computational system, RAS security is subject to deliberate cyberattacks, which can have disastrous consequences. One example is the well-publicised cyber-attack on an SUV Jeep in 2015 [5], which has triggered significant research among vehicle manufacturers in terms of determining and closing off vulnerabilities. As shown in Fig. 1, a red arrow is added from cyber space to physical space. Cyber threats to physical systems cannot be ignored. A trustworthy RAS should be impregnable to cyber-security attacks, protecting human personal data and health, while adhering to regulation compliance (e.g., GDPR).

*Internal System Health*. In addition to the challenges of safety and security, the reliability of RAS faces the normal challenges of any electronic/computational system, in terms of component failures, and unanticipated behaviours arising from less than 100% fault coverage. Faults in RAS occur due to erroneous sensor readings, incorrect interpretation of the (potentially incomplete) sensor data, or errors in actuator movement. The most difficult faults arise, in terms of the underpinning AI system, incorrectly interpreting sensor input or actuator output feedback, potentially leading to erroneous perceptions and disastrous decision making. Fault diagnosis is not a new problem and the electronics industry has been addressing it for decades, but its importance has escalated as systems become more complex and autonomous. It is paramount that reliability, graceful degradation and redundancy are core considerations in a RAS design, to ensure early detection of diverse potential abnormalities. It is crucial that the RAS design incorporates fault-tolerant operations to reduce performance degradation, ensuring pre-emptive notification of impending errors and thus reducing the potential for catastrophic or dangerous situations emerging [12].

*Human-Machine Interaction (HMI)* refers to the bidirectional interactions between the machine and the human user(s). HMI will occur via a user interface, based on the sensory modalities of humans – speech, touch, or vision, and potentially smell. A key consideration is whether the human is considered an equal partner in the decision loop, or alternatively has the right to override the machine decisions. The CPS model of NISP (Fig. 1) assumes human interaction in the decision process. The input of humans can be fed into the decision loop. In principle, this is correct but there raise two major issues: (a) the complexity of the decision-making process impacting the quality and pertinence of the human input and (b) the issue of interrupting a RAS mid-point in a critical process operation. In terms of being considered a trustworthy system, it is crucial that the human be able to meaningfully interact or interrupt the system. Such an interruption needs to be implemented in a manner that does not increase the danger or exacerbate a fault and to be based on efficient communication between humans and machine, involving the latest state of the art implementations of human machine interfaces. Such systems will incorporate advanced visualisation to allow the human to quickly assimilate and interpret the current and predicted states of the system, determine development trend or estimate statistics, and so maintain the consistency of systems in its (dynamically changing) operating environment. As a minimum, the "Design for Error" principles [13] should assume that faults will occur and provide appropriate scope for human interventions, where an "error" is broadly interpreted as any set of circumstances that causes the machine to deviate from the human-anticipated decision trajectory.

*Ethics* are of increasing concern to the public in implementing trustworthy autonomous systems. The moral standing and dignity of a human cannot be conveyed on any machine. While the vast majority of recent developments in RAS will improve the human condition and have positive impacts on society, their continued rapid development poses significant ethical issues. Many issues are well-rehearsed (e.g., accidents involving self-driving cars). Assume an extreme case, where an autonomous car has to decide whether to run over a group of schoolchildren or plunge off a cliff, killing its own occupants. From the viewpoint of a service provider, ensuring the safety of its users is the top consideration, but the more general public will have moral concerns regarding the schoolchildren. Kallioinen et al. [14] studied the ethical issues of autonomous vehicles through virtual experiments on some scenarios, such as child pedestrians vs. adult pedestrians; pedestrians on the road vs. pedestrians on the sidewalk; and car occupants vs. pedestrians. Their experimental results show that human drivers and self-driving cars were largely judged similarly. However, there was a stronger tendency to prefer self-driving cars to act in ways to minimize harm, compared to human drivers. When facing a non-deterministic problem, an intelligent system can be designed to make decision based on the probabilities of multiple states in terms of conventional rules. However, it is of vital importance to enable the final decision to follow the relevant regulation, rather than an engineer's preference, especially for an extreme case. The immediate problem is the speed of developments in AI and robotics, which far exceeds the rate at which the necessary accompanying ethical and moral framework is being developed. Therefore, common international standards, frameworks and guidelines need to be urgently agreed to inform RAS regulation and relevant laws.

### III. THE CHALLENGES IN IMPLEMENTING TRAS

The NIST CPS model can be applied to RAS, as most kinds of RAS belong to CPS, even if a fully autonomous system neither has a human intervention nor supports Internet connectivity. Namely, RAS complete their specific tasks through a close loop of four phases of CPS plus HMI. The optimization of RAS performance is facing great challenges brought by diversity, uncertainty and complexity of tasks for different application domains. The most challenge lies in that the performance and quality of RAS for various tasks in the context of a specific application domain must align with the five properties of a trustworthy RAS.

## A. Challenges to RAS Safety

Regarding the safety of RAS, two aspects need to be considered: the surrounding environment threats to the safety of RAS and RAS threats to the safety of users or assets in surrounding environments. They are mainly decided by perception technologies and the response behaviour of RAS. As shown in Fig. 1, the information, coming from the sensing system and representing the current state of the physical environment, is crucial, and it may also come from an operator. The decision phase receives the information from the sensing system or operators and generates action plans according to an abstract representation of the system and its environment. For example, in a robot's navigation system, the decision output is the optimal path on which the robot can avoid any obstacles and reach the destination. Hence, safe navigation requires the robot to be able to effectively and efficiently detect obstacles on the way towards the target location [15]. The decision functions mainly include learning, planning or goal reasoning, depending on the specific application domain. To enable RAS to work safely at various circumstances, the following challenges need to be tackled:

(1) Dynamic environments with uncertainty. RAS needs to be able to correctly sense their surrounding environments, deal with any exceptional cases or emergencies make real-time decision and produce a rapid response without deflecting the goal pre-defined for the RAS. For example, an autonomous car on road needs to deal with various exceptional cases, such as pedestrians change their mind suddenly to cross the road, road direction changes, an accident occurs ahead, or suddenly fog.

(2) Real-time synchronisation with other members when RAS is collaborating in a human/robot team. It requires a RAS to be able to safely complete various tasks, harmonically work with humans and other robots in a collaborative team and avoid any accidents that could harm other team members and itself.

(3) Diversity of unexpected failures. RAS needs to be able to detect unknown failures effectively and efficiently, tackle any exceptions, and provide timely warning and prompt response to any failures of the systems.

(4) Ability to safely perform in a situation for which RAS has no prior experience or pre-programmed response. A self-learning autonomous system may be able to deal with some unexpected events to reduce the risks that threaten the safety of the users, itself or others, but it cannot survive for all unexpected events that occur for the first time. Hence, at an emergent situation, timely human intervention at any execution point is of vital importance.

## B. Challenges of RAS Security

With the introduction of IoRT to tackle challenges from real-world applications, by using sensors, AI, software and communication technologies, etc., cyber threats move from IT infrastructure in the digital world to actuation systems in the physical world [16][17]. As a result, the attack surface of IoRT systems has been drastically enlarged. Cyberattacks or crimes not only threaten known devices in IT infrastructure at the upper layers of the IoT stack but also target conventional communication protocols and RAS in the lower level of the IoT stack. Due to the connectivity of RAS, more access points are potentially vulnerable to cyber attackers, and through these attack points, attackers can intrude a system, e.g., by injecting data to or extracting data from the system, thus compromising the security and control of the system.

Usually, researchers pay much attention on the cyber security of hardware and software of autonomous systems and communications between different devices [18]. However, there are rarely studies on cyber threats to developing platforms of RAS, which directly influence the security of the systems to be developed. Hence, the cyber threats to supply chain cannot be ignored [16].

The first challenge of RAS security is to secure communication links. Different applications may use different communication techniques. For example, public traffic tools (e.g., buses and trains) provide Wi-Fi services; GPS, radio and Bluetooth have been used in modern vehicles, and 5G communication between vehicles and road infrastructure is a future trend for connected and autonomous vehicles. However, different communication channels provide a means of intruding a RAS with various attack techniques, such as, general Trojan-horse attacks on Alice or Bob's system via the quantum channel [19], Man-in-the-Middle attack between vehicle-to-vehicle communication, MAC Spoofing, Wireless hijacking, Denial of Services (DoS), malicious eavesdropping [20], and attacks exploiting the vulnerable of Key Negotiation of Bluetooth (KNOB) [21].

The second challenge is to secure the integrity of RAS software. Integrity is one of the important security objectives in security triangles (confidentiality, integrity and availability). The integrity of system software requires protecting RAS software from code modification, malfunctions, loss of control and loss of personally identifiable information to loss of communication and network traffic congestion. Any broken integrity of software could cause severe consequences. Especially, the critical challenges lie in the loss of control and malfunctions, which could directly break the safety of RAS and even kill the users' lives of the RAS [22].

The third critical challenge is to secure hardware. An autonomous vehicle is composed of various components, including mechanical and electronic components, especially many Embedded Computing Units (ECUs), which could be targeted as an attack point. The resource limitation of embedded systems poses tight constraints on both communication and computing capacity. Hence, these constraints make it challenging to create advanced security solutions for embedded systems [23]. The hardware attack surface can be any possible components in RAS, such as sensors, USB ports or Input/output units, and embedded systems. If an attacker intrudes the sensing system of RAS, adversarial readings from sensors could cause a wrong decision, and further lead to a wrong action of the system. Typical side channel attacks on embedded systems for stealing secret information without a trace from the device, include fault injection attacks, power analysis attacks, time analysis attacks and electromagnetic analysis attacks [24]. For example, a Rowhammer fault injection attack can be mounted even remotely to gain full access to a device DRAM (dynamic random access memory); Cache side-channel attacks retrieve secret information by monitoring the cache of the system [25].

*C. Challenges to RAS Health*

In document ISO 10303–226, a fault is defined as an abnormal condition or defect at the component, equipment, or sub-systems level, which leads to a failure. Faults can be categorised into three types: deterioration (fatigue), sudden fault (noise) and initial failure [26]. Diagnosis is the ability to detect a fault, isolate and identify which component has a failure and analyze the potential impact of the failed component on the health of the system, whereas the prognosis is the capability to predict upcoming states of a system or a fault before it occurs [27]. RAS has higher requirements for fault prediction than other systems, as a fault in an autonomous system without human intervention could produce more severe consequences. However, it is challenging to obtain highly accurate prognostic information, as it greatly depends on the usage of the system, the experience of operators as well as the working environments. All of these uncertain factors directly influence the fatigue level and speed of system components, but it is difficult to quantify the measures of these factors.

RAS is a complicated system. For example, a vehicle has a complex mechatronic structure consisting of subsystems, such as gearbox, engine, brakes, fuel, ignition, exhaust, and cooling. Normally any subsystem comprises electromechanical processes, actuators and sensors. The sensors and actuators are associated and controlled with an engine control unit, which manages and screens the procedure. However, to save costs, a manufacturer may install low quality or too few sensors to monitor the system. Hence, the limited data from sensors may hamper the performance and coverage of diagnosis. Furthermore, another critical challenge is to meet the requirement of real time online diagnostics and prognostics.

The self-diagnosis system of RAS mainly includes three functions: internal state sensing, fault diagnosis & tolerance, and fast responses for sudden faults [26]. As sensor faults produce wrong data, which cannot reflect the state of the object to be monitored, the system could make an incorrect decision; Actuator faults can result in a wrong behaviour of the system; and faults of electronic components can cause malfunction and disorder of the system. The complexity, diversity and uncertainty of faults bring many challenges. For complicated faults, it is particularly challenging to find the fault origin, which in turn brings the challenge to implement fast and right responses.

*D. Challenges of Human-RAS Interaction*

No matter at which autonomy level a system is implemented, human-machine interaction (HMI) is indispensable in the system. Even if it is a fully autonomous system, HMI is still of essence in the system, as humans need to get the control of the system in an emergency. This is consistent with the modern view of human augmented by AI, as opposed to being replaced by AI and RAS. The studies of Veloso [28] proved that it is difficult to interrupt a robot during its autonomous execution, if we do not predesign an appropriate interruption mechanism, unless the power is shutdown. However, an unexpected power shutdown may damage the system. The design of allowing a human to interrupt a robot at any execution points is complicated. This requires us to traverse all scenarios, where different temporal-spatial parameters and constraints for different tasks, such as task priorities, operations, interruption frequency, and timings make it challenging to implement effective and efficient HMI.

Many applications require human and robots work collaboratively. For some safety-critical domains (e.g., defence, healthcare, and aerospace), the consequences of mis-operations, mis-responses and failures might be extremely bad and could make huge economic loss and even cause tragic accidents with the loss of human lives. One of the advantages of using RAS is that RAS could help humans complete diversity of dangerous, difficult or repeatable tasks. For example, RAS has played an important role in extreme and hazard environments. Such physical environments are dynamic, uncertain and even probably unknown. Hence, it is critically challenging to implement trusted Human-Robot Interaction in such a complex physical world.

Effective communications between humans and robots require robots to have the capacity to share their socially acceptable responses and common-sense knowledge to handle a broad variety of situations with clear interpretation and understanding of their complicated semantics. A critical challenge lies in diversity, complexity, and uncertainty of human status. For example, a social robot should express, understand and induce verbal and non-verbal emotions as part of the interaction process. Effective recognition of human emotions is critically challenging, as various human emotions are complicated and uncertain, and even single expressions represented by different subjects could be different very much.

Another key challenge for trusted HMI lies in the Theory of Mind (ToM), which is the capability of people to infer the intention and belief of others [29]. Namely, to ensure trusted HMI, we should improve the understanding of the mind of the two entities that communicate with each other. The robots' understanding of human ToM in intention, knowledge and competence can enhance the quality of the trustworthy interaction [30]. Currently, some computational ToM models have been proposed to enhance the robot's understanding of human user's intention and trustworthiness [31,32].

The inability to exhaustively test a complicated HMI system brings the most critical challenge to system verification. This may make users face many unexpected situations in use of these RAS that are tested incompletely. Hence, RAS should have the capacity of learning from, adaptation and response to unforeseen circumstances in a dynamic and changing world. Implementing such capacity of RAS will be a key step to go forward in the future.

*E. Challenges of Implementing RAS Ethics*

Ethics are moral principles that govern a person's behaviour or the conduct of an activity. It involves systematizing, defending, and recommending concepts of right and wrong behaviour [33]. RAS ethics are to address how human developers, manufactures and operators behave in order to minimise the ethical harms that RAS could produce, due to unethical design or misplaced applications. Some professional organizations have actively begun developing recommendations and policy statements. For example, IEEE published the document of "Ethically Aligned Design" to promote public understanding of the importance of addressing ethical considerations in the design of autonomous and intelligent systems [34], and the European Group on Ethics in Science and New Technologies issued a call for a "Shared Ethical Framework for Artificial Intelligence, Robotics and Autonomous Systems" in March 2018 [35]. Researchers from the UK-RAS network have identified seven ethics issues in RAS, i.e., Bias, Deception, Employment, Opacity, Safety, Oversight and Privacy [36], of which, Safety and Privacy have been addressed in the two properties of Safety and Security of RAS in this research. Bossmann [37] identified top nine ethics concerns associated with AI, such as Unemployment, Inequality, Humanity, Artificial Stupidity, Racist Robots, Security, Evil Genies, Singularity and Robot Rights. The European Parliamentary Research Service (EPRS) divided AI ethics into three phases of concerns: immediate, here-and-now concerns (e.g., data privacy and bias), near-and-medium term concerns (e.g., impact of AI and robots on jobs and workplaces) and longer-term concerns (e.g., possibility of superintelligence) [38]. Hence, the challenges of RAS ethics can be summarized in three phases with respect to a timeline.

(1) *Immediate concerns* are to put ethics into the design of an autonomous system. The first key challenge lies in the bias from designers, manufacturers, operators, and in particular, ML algorithms. Gerdes and Thornton [39] investigated the implementation of ethics in autonomous vehicles, concerning ethics with the constraints or cost in the design. A critical challenge is to deal with the trolley problem. There may be some cases that an autonomous system cannot simultaneously satisfy all of the constraints, but it must make a decision as to the best course of action. Similarly, an autonomous vehicle may need to be allowed to cross into an adjacent lane and drive against the flow of traffic if this would avoid an accident with another vehicle. Moreover, human prejudices in for example the choice of learning data may result in ethnic bias in the output of automated selection processes. The second key challenge lies in deception. Boden et al. [40] stated that robots are manufactured artefacts, and they should not be designed in a deceptive way to exploit vulnerable users. Further, the third key challenge is to implement transparency in the multi-facets' decision making of RAS for scrutiny and avoiding oversight. The fourth key challenge is the creation of regulation and laws that are accepted by public and applied to shape the behaviour of designers, manufacturers and operators to ensure the implementation of RAS ethics and avoid the production of evil genies.

(2) *Near-and-medium-term concerns* are about the roles of robots in the society. With the deployment of RAS, many jobs of humans can be done by robots. Humans will face the challenges – How to create a room to assume advanced roles for humans and how to shape the hierarchy of labour forces in the society. Robots' right is another challenge to be tackled. The fact that robot Sophia became a citizen of Saudi Arabia in 2017 raised much public attention. It was world's first AI humanoid robot who received a citizenship from a country [41]. However, we have not seen any regulation or laws to clearly formulate the right of robots.

(3) *The longer-term concern* is about the possibility of robots reaching or exceeding human capacities (so called superintelligence) [38]. The challenge lies in how to govern the innovation of RAS and avoid superintelligence and singularity, at which, technological growth becomes uncontrollable and irreversible, resulting in unforeseeable changes to human civilization.

These three-phases of concerns are calling for human-centered AI for the development of trustworthy RAS, which will be addressed in Section V.

## IV. OPPORTUNITIES OF AI TECHNOLOGIES

AI technologies have been applied in widespread areas, such as extreme environments, social-health care, manufacturing and military. The success of AI applications for various purposes has demonstrated the key role of AI in implementing trustworthy RAS. For example, to ensure the safety and reliability of robots with a high degree of operational autonomy under uncertain conditions, Zhao, et al. [42] developed a Bayesian inference and imprecise probability model on the layered Markov model, verified by the case of unmanned underwater vehicles in extreme environments. The research of Zhou and Yang [43] shows deep convolutional neural network for 2D biomedical semantic segmentation out-performs conventional methods in both accuracy and levels of automation. Lee et al. [44] surveyed the state of AI technologies and their power in ecosystems to implement the requirements of industry 4.0. Moreover, AI based trajectory and payload optimization of Rover are key steps to the Mars mission of NASA in 2020 and the landing plan in February 2021 [45].

To improve the trustworthiness of RAS, the performance and quality of services provided by RAS need to be complemented with the trust properties. Namely, RAS should be robust for any system health issues, safe for any uncertain and dynamic environments, secure for any cyber threats, cyberattacks and cybercrimes, and tolerant for any mis-operations by users and allow users to intervene at any execution point, even if it is a fully autonomous system.

### A. AI for the Safety of RAS

A monitoring system is essential to ensure the safety of RAS [1]. NASA updated a conventional monitoring system architecture by including user inputs in the monitoring loop, as shown in Fig. 2.

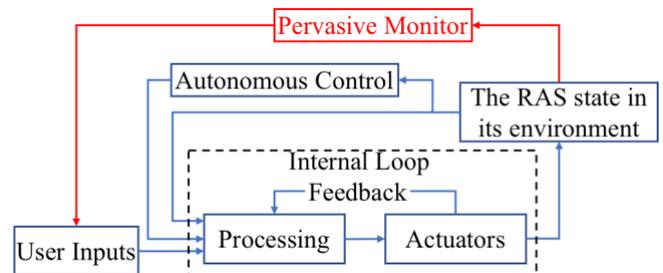

Fig. 2. Pervasive Monitoring Architecture, derived from [1]

Safety critical systems, including various types of RAS, such as UAV, aeroplanes, autonomous vehicles, using a Safety

Instrumented System (SIS) with specific control functions to maintain safe operations of their processes when unacceptable or dangerous conditions occur. Many techniques can be used to implement SIS for various types of autonomous systems. A sensing system should have the functions of data collection, data pre-processing and perception, etc. A key function of SIS is to recognize anomalies in their surrounding environments. Advanced sensing techniques are essential to ensure the performance of SIS. The most frequently used sensors include laser sensors (LIDAR) [46], visual sensors [47], radar, GPS, infra-red sensors [48] and ultrasonic sensors [49].

Aviation systems have the highest standard for safety in all autonomous systems. There are many monitoring subsystems in an aircraft, such as instrument monitoring, system monitoring, and environment monitoring. Autonomous navigation is an important challenge in RAS. It is to enable a RAS to run safely and autonomously within the surrounding environments, which might be uncertain and dynamic. For example, the capability of UAV navigation relies on advanced sensing systems and intelligent control algorithms, which refer to its running status and spatial information from all monitoring systems and adjust its flying behaviour appropriately.

Various AI techniques, especially ML techniques, have been applied for the perception of the surrounding environment and driving the navigation of RAS. For example, Ouarda [50] proposed a neural path planning approach for mobile robots; He et al. [51] developed a linguistic decision tree model to solve the classic robot routing problem through decomposing a robot's task or behaviour to a few atomic units; Huang et al. [52] developed dynamic obstacle detection system with a support vector machine using the space-time feature vector from Lidar for driverless cars; Wu et al. [53] transformed the task of path planning into a task of environment classification, for which a deep convolutional neural network (DCNN) was performed to guide the direction of robots; Zhu et al. [54] proposed a two-stages speed sign recognition system for autonomous vehicles in dynamic environments, through a Spatial Pyramid Pooling DCNN for salient target detection based on the background absorbing Markov chain; Mohanta and Keshari [55] developed a two-stages path planning by using a probabilistic roadmap method to generate the shortest path between the start position to target position in a priori known cluttered environment, with a head angle adjustment to ensure the smoothy turning throughout the path.

However, there still lacks of an effective mechanism or architecture of implementing efficient online ML that can adapt to a dynamic environment. Also, improving the perception (e.g., obstacle detection), positioning accuracy, decision accuracy, and resilience to the surrounding environment are still serious challenges for RAS.

*B. AI for RAS Security*

RAS operates in real-time. Human power is often not quick enough to protect RAS. Nevertheless, security automation may be necessary to effectively and efficiently secure RAS, greatly mitigate the risk from cyber threats, and enable prompt response for any cyber incidents caused by cyberattacks and cybercrimes, thus reducing the impact of cyber incidents on RAS and economic loss. AI technology is indispensable to the creation of the security automation for RAS. Unlike other problem domains, cyber security of RAS is to resist the adverse behaviour of determined and sophisticated attackers. RAS usually are connected to the Internet for improving their computing capacity, storage capacity, accessibility, usability and flexibility, etc, but they could be the target of hackers for different purposes. Cyber Intelligence is expected to be able to secure the benefits to all from the cyber-connected world. An architecture for RAS with security automation is demanded, and it should enable the implementation of "Security by Design", required by Industry 4.0, with the compliance of adaptive, self-learning and autonomous security [56]. AI techniques have been applied for cyber security in two aspects:

For access control, pattern recognition techniques have shown the power for bio-metrics authentication (e.g., fingerprint, face, iris and palm), signature and keystroke verification. For example, Fang et al. [57] proposed a fast & holistic authentication and authorization approach to analyzing the complex dynamic environment through online ML and trust management, thus achieving adaptive access control. User Entity Behaviour Analytics (UEBA), a new proactive approach to security, is a type of security process that uses ML algorithms and statistical analysis to detect real-time network attacks [58].

ML techniques have played important roles in data-driven cyber security, as they bring two significant gains to threat intelligence: first, machines can deal with huge amounts of data and their complex relations, which are impossible to do by humans; second, machines can implement the automation of cyber security, which it is not possible to implement by humans. There has been much research in this area, such as intrusion inspection [59], anomaly identification [60], web robot detection [61], and malware recognition [62].

The ongoing goal of ML-based intrusion detection systems (IDS) is to improve the accuracy and reduce the false alarm rate of detecting unknown attacks. Due to the constant development of attack techniques, first, an IDS should passively operate at the network level to detect intrusion, thus reducing the impact of cyber-attacks on RAS; second, IDS should be adaptive to be able to detect new attacks. To secure RAS enabled by IoRT, intrusion detection in edge devices is necessary. However, the constraints of computing capacity and critical requirement in real-time performance on edge devices may limit the capacity of edge IDS. Hence, creating effective and efficient edge IDS is a critical challenge. ML algorithms have been developed to identify DoS attacks to secure IoT enabled systems [63]. With the growing number of cyber-attacks and increasingly complex IT environments, an intelligent incident response mechanism is more than just a set of instructions. The automation is the best approach to empowering a fast incident response [64]. A socio-technical model-based effective and efficient threat intelligence with alerting enrichment and a priority order of actions on RAS could be a feasible solution for securing IoRT enabled systems.

However, as mentioned in [65], ML models, as the key driver of cognitive cyber security, may be hacked, as the implementation of ML algorithms is a programme or a function in a programme. The adversary behaviour of hackers could not only modify the code, but also insert or replace training samples with adversary samples, which is an instance with small, intentional feature perturbations that cause a ML model to make a false prediction. Adversarial examples make ML models vulnerable to attacks. The consequence of an adversary sample could be severe for RAS and directly damage the trustworthiness of RAS. For example, a driverless car could crash into another car or pedestrians, because it ignores a stop sign, when a hacker had placed a picture over the stop sign.

Therefore, the protection of ML models should overcome two challenges: one is to secure the code of ML models, which include training code and test code, and the other is to solve the adversary-fitting problem. For the first challenge, the effort is the same as on the protection of general software systems; In real-world applications, it is difficult to check if the collected data is wrong or not. Some classic adversarial defence techniques, described in [66], include

(1) Adversarial training, an intuitive defence method against adversarial samples, which attempts to improve the robustness of a neural network by training it with adversarial samples.
(2) Randomisation schemes for mitigating the effects of adversarial perturbations in the input/feature domain on Deep Neural Networks (DNNs), which are robust to random perturbations.
(3) Denoising (e.g., GAN-based input cleansing), a straightforward method for mitigating adversarial perturbations/effects. There are two ways of denoising (i) partially or fully remove the adversarial perturbations from the inputs, and (ii) alleviate the effects of adversarial perturbations on high-level features learned by DNNs.
(4) Provable defence techniques based on well-defined types of attacks.

ML, as a key technology, is driving the development of security automation, but we must ensure that correct (or trustworthy) features or data are fed into ML models for their own security. Hence, the robustness and security of data collection systems and ML models need to be investigated in the system design.

No matter what kind of ML technique is used to detect cyber-attacks or anomalies, the key performance indicators include accuracy, F-measure, Confusion Matrix, ROC curves, Mean-Squared Error, Standard Deviation, etc. While improving the true positive rate is the goal for anomaly detection, one cannot ignore the false positive rate. Especially, in applications of RAS, real-time performance is strictly required, even if the computing capability of RAS may be limited for on-board countermeasures to secure RAS. This brings critical challenges in the implementation of security automation.

*C. AI for the Health of RAS*

One of the important factors that affect the reliability of a system is the health of the system and the capacity of self-diagnosis. Fault diagnosis is an action of identifying a malfunctioning system based on observing its behaviour. It can be an important technique to ensure the safety of RAS. With the development of AI techniques, many new methods have been applied for fault diagnosis. Dynamic artificial immune system is one of the AI methodologies with strong ability of self-learning and self-adaptability.

Since the 1980s, *Analytical redundancy* methods have been a main trend for fault diagnosis [67]. Fig. 3 illustrates the framework of an analytical fault diagnosis model, where, $f_a$ is an actuator fault, $f_c$ is a process/component fault, and $f_s$ is a sensor fault.

A fault diagnosis algorithm is conducted to check the consistency of the feature information of the real-time process carried by an input $u$ and the output $y$ against the pre-knowledge on a healthy system, and a diagnostic decision is then made by the diagnostic logic. Now, with the development of AI or ML techniques, diagnostic algorithms can be implemented using a ML model, which can be trained by a set of historical data from the status information of sensors, actuators and components, including fault information, $f_a$, $f_c$ and $f_s$.

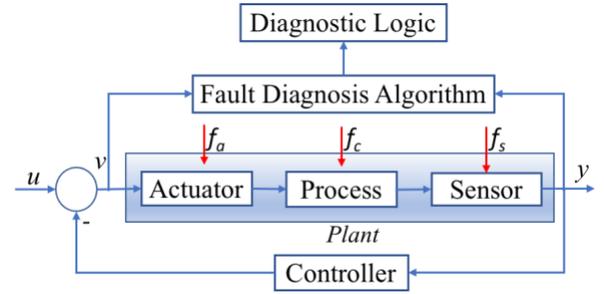

Fig. 3. Analytical Fault Diagnosis [12]

Usually, fault diagnosis has three tasks, such as fault detection, fault isolation, and fault identification. As the most basic task of the fault diagnosis, fault detection is used to check whether there is a malfunction or a fault in the system and determine the time when the fault occurred, and it can be transferred to a decision-making problem in the ML domain. The aim of fault isolation is to determine the location of the faulty component, and it could be transferred to an optimization problem. Fault identification is to differentiate the classes, levels and dimensions of a fault, and it can be solved with ML techniques as a classification problem. From the technical aspect, fault diagnosis technology can be divided into four categories: model-based fault diagnosis, signal-based fault diagnosis, knowledge-based and hybrid methods.

Model-based diagnosis is the activity of locating malfunctioning components of a system solely on the basis of its structure and behaviour. It performs various diagnostic tasks via on-line reasoning and inference of a system's global behaviour from the automatic combination of local models of its components. Obviously, ML techniques could be used to create models that map the relationships between inputs and outputs (Eq. (1)), where $v$ is the sum of input $u$ and the feedback from the controller in Fig. 3.

$$\hat{y} = \mathcal{L}(f(v)) \quad (1)$$

The system will check that the output $y$ is consistent with the model output $\hat{y}$; if they are inconsistent, the system is considered faulty. For example, Hashimoto et al. [68] developed a multi-model approach to identifying three failure modes (e. g. hard failure, noise failure, and scale failure) of faults in the five internal sensors on a mobile robot based on Kalman filters.

Instead of explicit input–output models, signal-based fault diagnosis usually uses three types of measured signals: time domain, frequency domain and time-frequency domain to determine faults [12]. As the measured signals reflect faults in the process, we can extract the features from measured signals, and use symptom analysis and prior knowledge on the symptoms of the healthy systems to make a diagnostic decision. To implement automatic fault diagnosis, various ML techniques can be used with the inputs of measured signals, features extracted from signals or raw data from sensors. For example, Eski et al. [69] used artificial neural networks to predict faults based on the noise and vibration of robot's joints; Cho et al. [70] utilised neural network to estimate the fault torque of adaptive actuator for robot manipulators; Ran et al. [71] categorised three types of fault diagnosis and prognosis in predictive maintenance systems, such as knowledge based, traditional ML based and Deep Learning based approaches.

Knowledge-based fault diagnosis methods can effectively use expert knowledge and experience to make judgements [72]. Knowledge representation provides clues for ontology reasoning in fault diagnosis. However, in the actual fault diagnosis process, it may be difficult to determine the relationship between the fault phenomenon of the components in RAS and the cause of the fault. A fault phenomenon on a component could have many causes, while a fault could show various types of phenomenon on different components. Due to the complex structure of RAS, multi-source failure and the suddenness of a failure, the combination of empirical knowledge and mechanism principles could be used to solve various fault problems. While the fuzzy knowledge representation could provide an effective approach to representing the uncertainty of knowledge [73], a dynamic uncertain causality graph can be useful to illustrate the relationship between fault phenomenon and causes [74].

To improve the accuracy of fault diagnosis for RAS, a hybrid approach, combining various ML models and prior knowledge in problem space, can be a good solution. For example, Nadeer et al. [75] provided an online fault diagnosis scheme for a spark ignition automotive engine, using a hybrid model with three stages: a single extended Kalman Filter estimator, a residual prediction stage, and a fault detection and isolation stage.

Diagnostic accuracy, real-time performance, and data availability are key challenges for data-driven fault diagnoses, same as for data driven cyber security. Effective and efficient allocation of a fault root is a critical challenge for a complex system comprising many correlated components. Although AI optimization techniques can improve the accuracy of fault allocation, it could be difficult to reach the real-time performance.

A swarm robotic system can be a large-scaled distributed system, consisting of a group of robots (e.g., drones), working for a mission with swarm intelligence for extreme and hazardous environments or entertainment. Such a large swarming system without the need of human intervention requires employing autonomous self-diagnosis, self-healing and self-reproduction at certain circumstances, where human does not have efficient power to deal with. For example, Dai et al. [76] proposed a multi-functions model for a swarm robotic system, including virtual neurones, running in a robot as a background process for perception and reflex, autonomous self-diagnosis via consequence-oriented prescription, autonomous self-curing, and self-reproduction.

*D. AI for Trusted Human-Machine Interaction*

Trusted Human-Machine Interaction (HMI) is a challenge for Human-centered Artificial Intelligence (HAI) [77]. HMI needs the harmonic collaboration of interdisciplinary areas. It involves human behaviour and mind modelling to enhance the capacity of robotic recognition; knowledge acquiring, representing, and manipulating at the human level; reasoning and decision making; thus, eventually instantiating physical actions both legible to and in coordination with humans. HMI researchers strive to leverage-advanced technologies from AI and quantum computing into easy-to-use human-machine interaction systems that align with our lives.

*Natural Language Processing* (NLP) is an important technique for improving the cognitive capacity of HMI, and the NPL capacity of RAS is an important indicator of human level machine intelligence [73]. For decades, classic shallow ML models (e.g., Support Vector Machine and logistic regression) have been used to solve NLP problems with high-dimensional and sparse features. In the last few years, deep neural networks based on dense vector representations have shown superior performance for various NLP tasks [78]. NLP mainly includes the two types of tasks: natural language understanding (NLU) and natural language generation (NLG). NLU includes the tasks of mapping the given input in natural language into useful representations, for example, a rule-based machine translation and of analyzing different aspects of the language. NLG involves text planning, sentence planning and text realisation.

*Robot Vision* uses the combination of camera hardware and computing algorithms to allow robots to process visual data from the world. It adds visual-understanding capabilities to a robot, so that the robot can perceive human non-verbal behaviour, and naturally interact with humans through body gestures, facial expressions and body poses. For example, a smart robot has been developed to assist physicians in performing surgery, using two cameras, a Near-Infra Red (NIR) camera from Basler and a panoptic 3D camera from Matrix to create high-contrast areas in the 2D NIR images [79]. Deep learning is a competitive technology for several computer vision benchmark problems, such as image classification, object detection and recognition, semantic segmentation, and action recognition [80].

*Tactile sensing* is a key technology to ensure that the physical human robot interaction is safe when such interactions in a shared workspace require physical contact between humans and robots. A tactile sensor can be used as the artificial sensitive skin of a robot, not only providing safety-related functions but also offering touch-based robot motion control, thus enhancing human-robot interaction [81]. Distributed tactile sensors can be easily used at different parts of the robot body. Their ability to estimate contact forces and to provide a tactile map with an accurate spatial resolution enables the robot to safely handle intentional touches and avoid unintentional collisions in safe human-robot collaboration tasks based on multi-sensor fusion [82].

*Knowledge extraction and sharing between humans and machines* is of the essence in the dynamic process of human-machine interaction. The knowledge from user's inputs and the outputs of HMI can be extracted by the machine. Artificial intelligence has made certain achievements in reflecting how the robot identifies and understands the external information, and thus performing the corresponding action. Various reasoning systems have been adopted for knowledge-based inference in human-machine interaction [83]. Tran et al. [84] has shown that a layerwise extraction can improve the performance of deep belief networks, and they proposed a symbolic characterisation approach for inserting prior knowledge and training of deep networks.

*Knowledge representation* can provide support for knowledge transformation to the user interface environment through modelling the abstractions of the knowledge [85]. It is important to create an adaptive knowledge representation process for automating HMI [86]. Devlin et al. [87] designed a model of Bidirectional Encoder Representations from Transformers (BERT) based on a deep *bidirectional* architecture to pre-train deep bidirectional representations from unlabelled text, which successfully support various NLP tasks.

Zadeh stated that fuzzy techniques can provide an effective approach to representing the imprecision and uncertainty of knowledge; fuzzy logic is a precise logic of imprecision and approximate reasoning, and a fuzzy set is a class with unsharp boundaries [73]. Hence, the recent developments in fuzzy technology can help make a system more robust.

## V. HUMAN-CENTERED AI FOR TRAS

There are many different concerns about human centered AI. In the context of TRAS, human centered AI is a means of improving the relationship between humans and RAS. Fridman [88] believes that implementing autonomous vehicles is the problem of integrating HMI, machine intelligence, psychology and policy, and he proposed seven principles in practice for human-centered autonomous vehicle: (1) share autonomy with humans, (2) learn from data, (3) human sensing, (4) share perception-control, (5) deep personalisation, (6) Imperfect by design and (7) system level experience. Nevertheless, human-centered AI for the implementation of TRAS may need to consider the following aspects.

### A. Acceptance Model of TRAS

Humans' acceptance of RAS decides the requirements for human-centered AI, enabling the implementation of TRAS. An acceptance model of trustworthy RAS is proposed with two aspects of worthiness and trustiness. Worthiness represents the quality of being good enough or suitability, trustiness represents a quality of being loyal or reliable, and adoption is the action or fact of choosing to take up, follow, or use something. Only if both worthiness and trustiness are designed into RAS, humans would accept and adopt it. As shown in Fig.4, while the functionality/ performance of RAS represents the worthiness of RAS, all the five properties and reliability of RAS represent the trustiness of RAS. The four properties in safety, security, system health and human-machine interaction are directly related to the reliability of RAS. Many models have been developed to explain human behaviour and their acceptance of new technologies. Humans may not be aware of the importance of cyber security. Hence, Zhang et al. [89] did not put security as an important element in the acceptance model of Connected and Autonomous Vehicles. He et al. [90] argued that security can directly affect the reliability and privacy, evidenced by a Denial of Service (DoS) attack, which utilises a flood of arbitrary packets to a target system and could cause a system malfunction, thus leading, for example, to a stuck or inoperable throttle, and raising potential safety concerns [91]. Therefore, cyber security is directly related to safety, but indirectly related to trustiness. Privacy depends on the protection of cyber security, but is directly related to the trustiness of RAS. The health of RAS could affect the human-machine interaction and fault tolerance should reflect the requirements of ethics, but is indirectly linked to trustiness. The reliability of RAS could affect the functionality and performance of RAS. Ethics is a key property, directly related to the trustiness of RAS. As previously stated in Section III, ethical issues include safety, security and privacy, etc. Hence, they are directly related to ethics. HMI may directly affect the safety. The dynamics of technology for HMI requires the development of innovative approaches to HMI, and methods to support the design of complex socio-technical systems under existing framework of ethics and regulations [92]. Hence, HMI is directly linked to Ethics.

In Fig. 4, solid lines represent direct relations and dashed lines represent indirect relations. Obviously, the proposed acceptance model is applicable for autonomous vehicles. Trustworthy autonomous vehicles, which satisfy the requirements under the framework of the acceptance model, would be adopted. Different application domains may have different weights of relations between different blocks in the model. For example, domestic robots may have less requirement in safety to the trustiness.

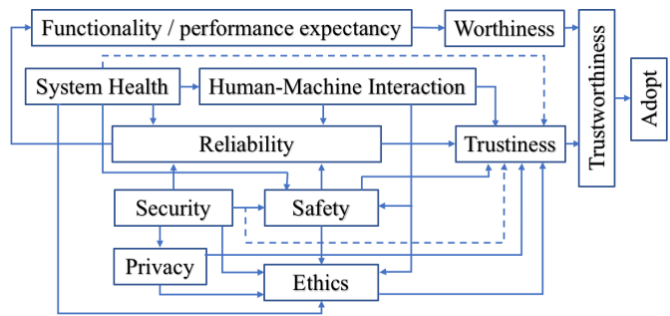

Fig. 4. Acceptance model for trustworthy RAS.

### B. Towards Human Level Intelligence

In agreement with PwC [93], we can briefly divide robotic intelligence into three levels: assisted intelligence, where AI has replaced many of the repetitive and standardized tasks done by humans (e.g., manufacturing machines); augmented intelligence, where humans and machines learn from each other and redefine the breadth and depth of what they do together (e.g., surgery robots); and autonomous intelligence, where adaptive and continuous machines take over the whole process of perception, decision making and actions, independent of human intervention (e.g., fully autonomous vehicles). Autonomous Intelligence is the most advanced intelligence. However, the autonomy of machines should be limited from the human's perspective.

As Zadeh stated in [73], humans have many remarkable capabilities, two of which stand out in importance: (1) the capability to reason, converse and make rational decisions in an environment of imprecision, uncertainty, incompleteness of information, partiality of truth and possibility, (2) the capability to perform a wide variety of physical and mental tasks without any measurements and any computations. There has been much evidence that shows machine intelligence has achieved a certain progress in the first capacity, in respects of linguistic intelligence, visual intelligence, hearing intelligence, tactile intelligence, spatial intelligence, emotional intelligence, (e.g., humanoid and programmable robot Nao [94]). However, the second capacity is still a long way off, as this requires a machine brain to have the capacity of strong perception of complex environments and prompt decision-making. With the first capacity, robots have been able to perceive simple environments or events. How to comprehensively and efficiently integrate such simple individual perceptions to form an embryonic, responsive nervous system for a robot is still unclear.

Clearly, the achievement of human level machine intelligence is still a challenge that is difficult to meet. Coinciding with the path towards a fully autonomous system, machine intelligence is on the way towards human level intelligence too. Now, the dimension of environments increases with the additional cyber space to the physical execution environments within systems and outside the systems. A prerequisite to the achievement of human level machine intelligence is the mechanisation of these capabilities [73]. In other words, a prerequisite of implementing TRAS is the automation of the four capacities of RAS aligning with the automation of RAS' functionality concerning different application domains. To create general human-level intelligence through integrating the automation of all RAS properties is not just a critical challenge, but also it is massive work, requiring many more generations of research and innovation. The human acceptance model to RAS may provide a framework of requirements to implement trustworthy RAS with human level intelligence.

### C. Augmented Human Capabilities

An important purpose of developing human-centered AI to implement TRAS is to augment human capacities and capabilities. RAS should effectively communicate and collaborate with humans, where each partner brings its own superior skills to the partnership — instead of seeking AI that will supersede humans but using AI to augment humans and human intelligence. In such a way, human-centered intelligent systems and intelligent human-machine interfaces could either amplify existing or create new human skills and capabilities. Such augmented humans and intelligence will allow humans to go beyond current human capabilities and offer new experiences [95]. Automatic machines have run in factories for a long time, doing highly repetitive and physically demanding tasks more efficiently and productively. This is called 'automation' intelligence, of which the automotive industry has demonstrated a good example. It was predicted that 6.6 million jobs across the ASEAN-6 economies could be made redundant by 2028 as a result of new technology adoption in terms of the report of Oxford Economics [96]. This requires entire business processes to be transformed and the jobs performed by humans to be redefined, much like the bank clerk's job was redefined with the advent of automated teller machines [95].

### D. Focus on AI's Impact on Humans

Technology is trusted if it benefits humans, is demonstrably fair, safe and reliable, is well regulated, and can be investigated if errors occur. Obviously, RAS has not reached this stage yet. Trust and justified confidence can accelerate technology adoption and job creation, and prevent a backlash against RAS, but only if trust is not misplaced [10].

Robots may take over many human activities. However, this does not imply that robots can replace and control humans. To develop trustworthy RAS, we must understand how artificial intelligence performs in practise and impacts on humanity. This requires humans to be cautious in the development of fully autonomous systems and follow regulations to avoid uncontrollable superior intelligence. While designing RAS, we should have a solution that allows humans to interrupt robots' work at any execution points and restrict the situation to one that would not harm humans. Namely, humans are guaranteed to be able to control the robots and avoid any risk that robots harm humans. This aligns with the first principle of sharing autonomy with humans in the loop, proposed in [88].

Much AI technology, is already in place or yet to be developed, can be used for both commercial and military applications. The biggest threat from AI is the potential for its weaponisation [97]. Contrary to this, we should foster the development of AI-enabled RAS in the areas, where RAS could (1) help humans work in extreme & hazard environments, (2) improve the capacity of health and social care, (3) augment the capability of manufacturing and food production, (4) reduce damage to the earth, (5) help recover the damage we have done to the planet, and (6) explore space. All of these global challenges and others should give rise to AI-driven mission-based innovations and should bring together citizens, scientists and engineers to address them. The AI for Good Global Summit in June 2017 discussed how AI could follow a development course able to assist the achievement of the United Nations' Sustainable Development Goals [98]. An AI partnership, comprised of more than 100 industries, including those large companies, such as Google, Microsoft and IBM, etc., has been established to get together diverse, global voices to realise the promise of AI [99].

Nevertheless, obeying Asimov's three laws of robots [100] is the essential condition:
(1) No robot may injure a human being or, through inaction, allow a human being to come to harm.
(2) A robot must obey the orders given it by human beings except where such orders would conflict with the first law.

(3) A robot must protect its own existence as long as such protection does not conflict with the first or second Law.

## VI. Conclusions

This research provides a comprehensive review on human-centred AI for trustworthy RAS. The contribution of the research lies in that (1) the five key properties of trustworthy RAS were identified with respects to safety, cyber security, system health, human-machine interaction and ethics; (2) the challenges in implementing trustworthy RAS were analytically assessed in respects of the five properties; (3) the roles of AI for implementing trustworthy RAS were explored in respects of safety, security, health and HMI, which need to reflect the requirements of ethics in the design of RAS, and (4) a structural approach was provided to RAS design.

While the performance and functionality of RAS represent important aspects of RAS, the five properties discussed ensure the trustiness of RAS. A new acceptance model of RAS is proposed to ensure both worthiness and trustiness of RAS for human acceptance and adoption to RAS. It provides a framework of requirements for implementing trustworthy RAS with human-centered AI. The new concept of human-centered AI (HAI) promotes the cooperation between technological innovation and humanistic and ethical consideration with three objectives: (1) to promote the technical frontier towards human intelligence; (2) to augment human's capabilities; and (3) to focus on AI's beneficial impact on humanity. Finally, it is highlighted that trustworthy RAS must, as a minimum, obey Asimov's three laws of robotics.

[100] I. Asimov, "Runaround". *I, Robot* (The Isaac Asimov Collection ed.). New York City: Doubleday. 1950. p. 40. ISBN 978-0-385-42304-5.

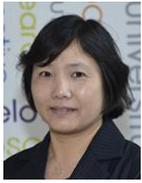
Dr Hongmei He (SIEEE'16, FHEA) is currently a Senior Lecturer at the School of Computer Science and Informatics at De Montfort University. She obtained her PhD degree in Computer Science from Loughborough University, UK in 2006. Previously, she worked as a lecturer or a postdoc researcher for several universities, such as Cranfield University, University of Kent, Ulster University and University of Bristol, for a wide range of AI applications, such as Cognitive Cybersecurity, Pattern Recognition, Cognitive Robotics, Computational Finance, Network-based Data Mining, Data/Sensor Fusion and Optimisations. Her current research focuses on AI for the safety and security of autonomous systems. She actively serves for IEEE UK & Ireland RAS Chapter as the chapter secretary and is the chair of the task force, "AI and Edge Computing for Trustworthy Robots and Autonomous Systems", in ADPRLTC of IEEE Computational Intelligence Society.

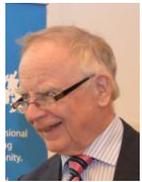
John Gray (MIEEE,FIET) is an Emeritus Professor of Robotics and Systems Engineering in the School of Electrical and Electronic Engineering at the University of Manchester. In 1988 he established the UK's National Advanced Robotics Centre, served as its Research Director and as a member of the managing industrial consortium ARRL. Prof Gray has subsequently been involved in a range of European Commission and industrial funded robotics research projects. In 2000 he was invited by Maff/Defra to establish and chair the Food Manufacturing Engineering Group (FMEG), which is an industrial forum to foster the uptake of automation. He has subsequently been involved in a large number of funded automation developments in this sector. He is currently an honorary editor of the Transactions on the Instruments of Measurement and Control and chair of the IEEE UK& Ireland section RAS Chapter.

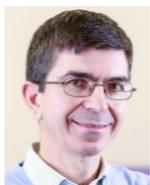
Angelo Cangelosi (SIEEE) is Professor of Machine Learning and Robotics at the University of Manchester (UK). He also is Turing Fellow at the Alan Turing Institute London, Visiting Professor at Hohai University and at Universita' Cattolica Milan, and Visiting Distinguished Fellow at AIST-AIRC Tokyo. His research interests are in developmental robotics, language grounding, human robot-interaction and trust, and robot companions for health and social care. His latest book "Cognitive Robotics" (MIT Press), coedited with Minoru Asada, will be published in 2021.

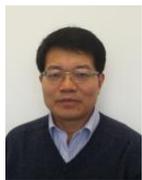
Qinggang Meng (SMIEEE) is currently a Professor in Robotics and AI with the Department of Computer Science, Loughborough University, UK. He is a fellow of the Higher Education Academy, UK. His research interests include biologically inspired learning algorithms and developmental robotics, service robotics, agricultural robotics, robot learning and adaptation, multi-UAV cooperation, human motion analysis and activity recognition, activity pattern detection, pattern recognition, artificial intelligence, computer vision, and embedded intelligence.

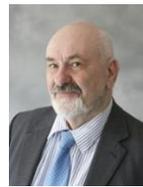
Prof T. Martin McGinnity (SMIEEE, FIET) received a First Class (Hons.) degree in Physics in 1975, and a Ph.D degree from the University of Durham, UK in 1979. He currently holds a part-time Professorship in both the Department of Computer Science at Nottingham Trent University (NTU), UK and the School of Computing, Engineering and Intelligent Systems at Ulster University. Before taking semi-retirement, he was formerly Pro Vice Chancellor and Head of the College of Science and Technology at NTU, Dean of the School of Science and Technology at NTU, Head of the School of Computing and Intelligent Systems at Ulster University; Professor of Intelligent Systems Engineering at Ulster University and Director of the Intelligent Systems Research Centre in Ulster University. He is the author or co-author of 350+ research papers and leads the Computational Neuroscience and Cognitive Robotics research group at NTU. His research interests are focused on artificial intelligence, computational neuroscience, modelling of biological information processing and cognitive robotics. His current projects are related to the development of biologically compatible computational models of human sensory systems, including auditory signal processing; human tactile emulation; human visual processing; sensory processing modalities in cognitive robotics; and implementation of neuromorphic systems on electronics hardware. His work finds applications in industrial robotics, data analytics and medical systems.

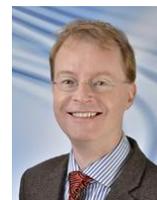
Prof. Jörn Mehnen (MIEEE) research interests at the University of Strathclyde, UK, focus on Industry 4.0 and Digital Manufacturing. He is concerned with the conversion of deep academic insights into industrially highly applicable knowledge, skills, and technologies. His research covers Trustworthy IIoT, AI and Robotics, Additive Manufacturing, Cloud Manufacturing, Mixed Reality and Digital Twins for Manufacturing.